\documentclass[conference]{IEEEtran}
\IEEEoverridecommandlockouts
\usepackage{amsmath,amssymb,amsfonts}
\usepackage{algorithmic}
\usepackage{graphicx}
\usepackage{textcomp}
\usepackage{xcolor}
\usepackage{hyperref}
\usepackage[square,sort,comma,numbers]{natbib}
\usepackage{soul}
\usepackage{float} 
\usepackage{enumitem}
\usepackage{booktabs}
\usepackage{stfloats} 
\usepackage{caption}     

\usepackage[inline]{trackchanges}
\usepackage[nameinlink,noabbrev,capitalize]{cleveref}

\usepackage{tikz}

\newcommand\copyrighttext{%
  \footnotesize \textcopyright 2025 IEEE. Personal use of this material is permitted.
  Permission from IEEE must be obtained for all other uses, in any current or future
  media, including reprinting/republishing this material for advertising or promotional
  purposes, creating new collective works, for resale or redistribution to servers or
  lists, or reuse of any copyrighted component of this work in other works.}
\newcommand\copyrightnotice{%
\begin{tikzpicture}[remember picture,overlay]
\node[anchor=south,yshift=10pt] at (current page.south) 
  {\fbox{\parbox{\dimexpr\textwidth-\fboxsep-\fboxrule\relax}{\copyrighttext}}};
\end{tikzpicture}%
}

\addeditor{NA}
\addeditor{PM}
\addeditor{SE}
\addeditor{AB}

\newif\ifhighlight
\highlighttrue 

\def\BibTeX{{\rm B\kern-.05em{\sc i\kern-.025em b}\kern-.08em
    T\kern-.1667em\lower.7ex\hbox{E}\kern-.125emX}}
\begin{document}
\title{Camera Control at the Edge with Language Models for Scene Understanding
}

\author{\IEEEauthorblockN{Alexiy Buynitsky}
\IEEEauthorblockA{
\textit{Purdue University}\\
West Lafayette, IN, USA \\
abuynits@purdue.edu}
\and
\IEEEauthorblockN{Sina Ehsani}
\IEEEauthorblockA{\textit{Armada AI} \\
Bellevue, WA, USA \\
se@armada.ai}
\and
\IEEEauthorblockN{Bhanu Pallakonda}
\IEEEauthorblockA{\textit{Armada AI} \\
Bellevue, WA, USA \\
bp@armada.ai}
\and
\IEEEauthorblockN{Pragyana Mishra}
\IEEEauthorblockA{\textit{Armada AI} \\
Bellevue, WA, USA\\
pm@armada.ai}
}

\maketitle
\copyrightnotice

\begin{abstract} 
In this paper, we present Optimized Prompt-based Unified System (OPUS), a framework that utilizes a Large Language Model (LLM) to control Pan-Tilt-Zoom (PTZ) cameras, providing contextual understanding of natural environments.
To achieve this goal, the OPUS system improves cost-effectiveness by generating keywords from a high-level camera control API and transferring knowledge from larger closed-source language models to smaller ones through Supervised Fine-Tuning (SFT) on synthetic data. This enables efficient edge deployment while maintaining performance comparable to larger models like GPT-4.
OPUS enhances environmental awareness by converting data from multiple cameras into textual descriptions for language models, eliminating the need for specialized sensory tokens. In benchmark testing, our approach significantly outperformed both traditional language model techniques and more complex prompting methods, achieving a 35\% improvement over advanced techniques and a 20\% higher task accuracy compared to closed-source models like Gemini Pro.
The system demonstrates OPUS's capability to simplify PTZ camera operations through an intuitive natural language interface. This approach eliminates the need for explicit programming and provides a conversational method for interacting with camera systems, representing a significant advancement in how users can control and utilize PTZ camera technology.

\end{abstract}

\begin{IEEEkeywords}
AI in Robotics, Large Language Models, Supervised Fine-Tuning, Human-Robot-Environment Interaction, Model Distillation
\end{IEEEkeywords}

\section{Introduction}

\begin{figure*}[ht!]
    \centering
    \includegraphics[width=1.01\linewidth]{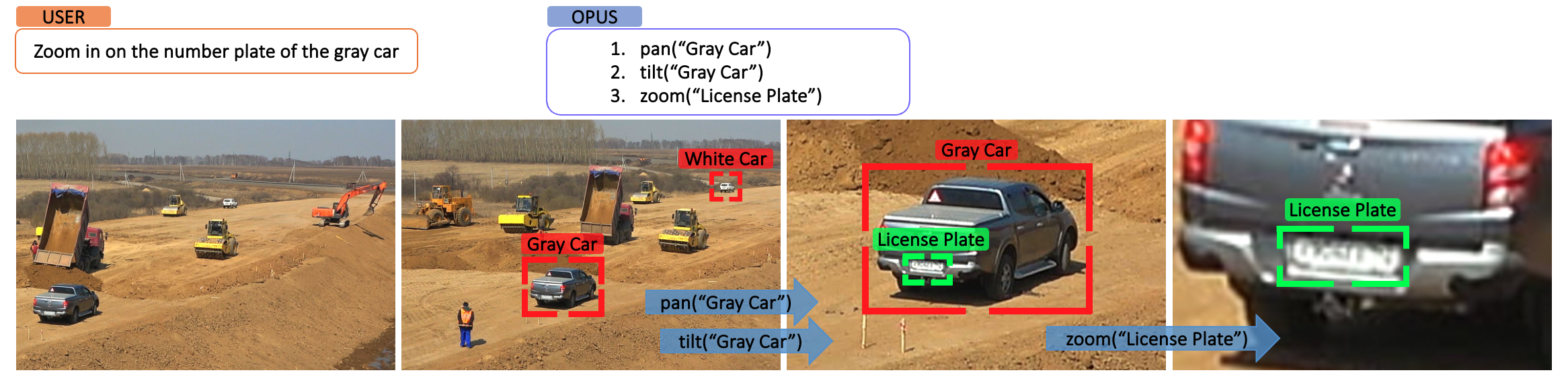}
    \caption{Consider a user-specified task such as "Zoom into the license plate of the gray car." The OPUS Interface processes the video frames, identifying objects in the scene—such as cars, excavators, dump trucks, rollers, and people—and converts them into text tokens. These tokens are passed to the LLM, which generates a sequence of high-level API commands to accomplish the task. The commands are then sent back to the OPUS Interface, which parses and executes them on the robotic system—in this case, the PTZ camera—performing the requested task seamlessly.}
    \label{fig:opus_intro}
\end{figure*}

Recent progress in Large Language Models (LLMs) has produced AI systems capable of logical reasoning, creativity, and problem-solving across diverse domains \citep{brown2020language, openai2024gpt4}. These models leverage the scale of their architectures and are trained on vast datasets, encompassing diverse sources—including books, websites, technical papers, and code repositories. This extensive training enables them to perform complex tasks beyond mere text generation or document summarization \citep{taori2023alpaca}, demonstrating abilities in code completion \citep{chen2021evaluating}, dialogue management \citep{xu2022long}, and complex reasoning \citep{yao2023tree}.

When integrated into robotics and industrial automation systems, this capability provides a pathway toward embodied intelligence-robots that can understand and interact with their physical environments in meaningful ways. For example:
\begin{itemize}[leftmargin=7pt]
 \item \textbf{Scene Understanding:} Recognizing objects, their attributes, and their relationships in a visual scene to make informed decisions.
 \item \textbf{Task Planning:} Generating and executing sequences of actions based on textual descriptions or commands.
 \item \textbf{Adaptive Interaction:} Modifying behavior dynamically based on changes in the environment or user input.
\end{itemize}

By combining LLMs’ reasoning and language-generation abilities with the sensory and actuation capabilities of robots, we can enable systems that not only process information but also act upon it in real-world settings. This integration has the potential to transform fields such as autonomous navigation, industrial automation, and assistive technologies, bridging the gap between abstract reasoning and physical embodiment.
Despite these promising capabilities, practical implementation remains challenging. Leveraging LLMs for robotic control poses significant hurdles, as it requires grounding the models in the robot’s operational environment while demanding low-latency processing on resource-constrained hardware, often without reliable network connectivity. Given these challenges, several approaches have been proposed to bridge the gap between abstract reasoning and real-world robotic control.

\begin{itemize}[leftmargin=0pt]
\item[] $(i)$ \textbf{Post-hoc Grounding with External Mechanisms}: 
Approaches that use LLMs as heuristics to guide additional policies—with grounding applied post hoc via external mechanisms \citep{ahn2022can, ma2023eureka} have been proposed. While effective in enabling creative solutions, these approaches can lead to mismatches between the suggested actions (or reward designs) and the robot’s actual capabilities, necessitating complex affordance checks and potentially leading to execution failures.
\item[] $(ii)$ \textbf{Embedding Sensory Inputs into LLMs}: Incorporating sensory observations directly into the model inputs, by embedding images or sensor data alongside language tokens \citep{driess2023palme, brohan2023rt1, brohan2023rt2}, thereby grounding the foundation models in the robot’s environment. However, this approach necessitates larger models and increased computational resources because of the high-dimensional nature of the sensory data, which limits deployment on resource-constrained devices and increases inference times. Additionally, these methods often struggle to generalize to unseen tasks.
\item[] $(iii)$ \textbf{Code Generation for Robotic Policies}: Several approaches have demonstrated promise in generalizing to open-ended tasks by grounding the robot using textual descriptions of its environment and observed objects \citep{vemprala2023chatgpt, mu2024robocodex, liang2023code, singh2023progprompt}. However, these methods rely on large-scale open-source models or closed-source models that cannot be deployed at the edge due to resource constraints. Furthermore, many code-generation approaches require constant human supervision to catch errors, thereby limiting their practicality for autonomous robotic control.

\item[] $(iv)$ \textbf{Multi-Step Prompting Systems with Multiple Planners}: Approaches that employ multi-step prompting systems with multiple planners to enhance execution precision \citep{lan2024llm4qa, shinn2023reflexion, singh2023progprompt} can indeed improve task execution by decomposing tasks into sub-tasks and incorporating feedback loops. However, these methods are complex, require an intricate setup, and are resource-intensive—resulting in longer inference times and potential scalability issues.
\end{itemize}

While these approaches advance integration between LLMs and robotic systems, each presents trade-offs regarding complexity, computational efficiency, and practical deployment, highlighting the need for a streamlined solution. 
Building on these insights, we introduce the \textit{Optimized Prompt-based Unified System (OPUS)}, addressing these limitations by directly integrating LLMs for efficient and accurate robotic control of Pan-Tilt-Zoom (PTZ) cameras. As illustrated in \Cref{fig:opus_intro}, given a task such as “zoom into the number plate of the gray car,” OPUS translates visual observations into concise textual descriptions (e.g., “a gray car is in the scene”). It then employs a small fine-tuned LLM to generate high-level, directly executable API commands, controlling the camera’s pan, tilt, and zoom functions. OPUS thus effectively mitigates the challenges seen in previous methods, including code-generation errors, high computational requirements, complex multi-step prompting, and mismatches between LLM outputs and robotic capabilities. Specifically, OPUS makes three primary contributions:

\begin{itemize}[leftmargin=0pt]
\item[]$(i)$ \textbf{Enhancing Efficiency and Deployability}: By converting the camera’s visual and sensory data into textual descriptions \citep{Jocher_Ultralytics_YOLO_2023}, OPUS grounds the LLM in the camera environment without the need to embed high-dimensional sensory data into a vision-language model. This approach reduces computational load, enabling deployment on resource-constrained devices.
\item[]$(ii)$ \textbf{Simplifying Control through Fine-Tuning}: OPUS leverages Supervised Fine-Tuning (SFT) of the LLM on a synthetic dataset tailored to PTZ camera tasks. This distillation enables a compact LLM to match the performance of larger, closed-source models while generating accurate and efficient command sequences in a single step—thereby avoiding the complexity and resource demands of multi-step prompting systems.
\item[]$(iii)$ \textbf{Eliminating Code-Generation Errors and Human Intervention}: OPUS employs a predefined set of high-level API commands for PTZ camera control, ensuring that the LLM outputs are directly executable and aligned with the camera’s capabilities. Unlike code-generation approaches that may produce erroneous code requiring oversight, our method minimizes errors and reduces the need for human intervention.
\end{itemize}

We validate our approach through experiments that demonstrate the effectiveness of the SFT-enabled LLM in PTZ camera applications. Our results show that OPUS performs favorably compared to baseline models and larger models such as GPT-4 (\Cref{sec:results}), confirming that our simplified approach achieves superior performance while addressing the key challenges of integrating LLMs into robotic control. Additionally, we present a PTZ camera simulation with comprehensive evaluation metrics and a test dataset of expert demonstrations, providing a valuable resource for benchmarking and future research.

The rest of the paper is organized as follows: Section II describes the OPUS framework in detail, Section III presents experimental results demonstrating our system's advantages, and Section IV concludes with implications and directions for future work. 


\section{Method}

\begin{figure}[ht!]
    \includegraphics[width=\linewidth]{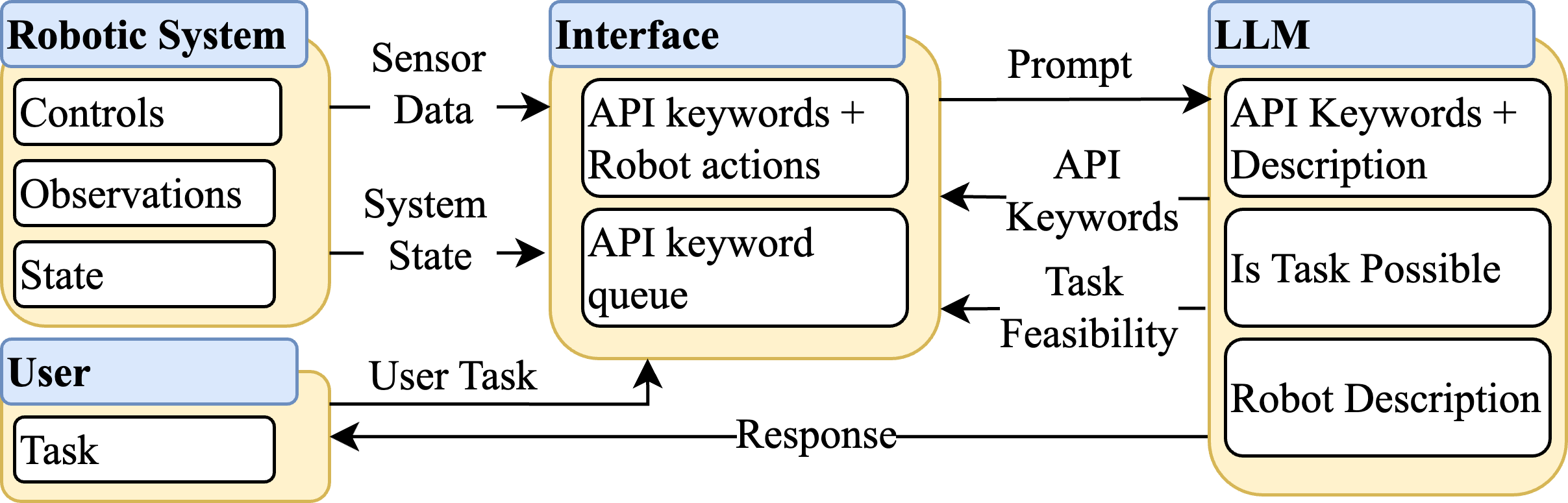}
    \caption{OPUS consists of four main components: Interface, LLM, User, and Robotic System. The LLM is responsible for reasoning through tasks, while the Interface converts non-text sensory observations from the Robotic System into text for the LLM. The Interface manages the parsing and execution of API keywords generated by the LLM.}
    \label{fig:opus_detailed_diagram}
\end{figure}

\subsection{Problem Statement and System Overview}

To ensure clarity, we introduce the following notation for our PTZ camera control system. Let $\prod = \{\pi_i\}_{i=1}^m$ represent the set of available robotic policies, where each policy $\pi_i$ corresponds to an executable robotic command (e.g., \verb|zoom(object)|). Associated with these policies is the set $T = \{t_i\}_{i=1}^m$, containing textual descriptions for each policy. Observations from the robotic system, denoted as $O$, include object labels and relevant environmental information derived from sensory data.

Given a user's request $R$, our model $M$ (typically an LLM) is tasked with mapping the textual descriptions $T$, environmental observations $O$, and the user's request $R$ into an ordered set of executable robotic policies:

\begin{equation}
\textstyle
M(T,O,R) = \{\pi_i\}_{i=1}^n, \text{ where each } \pi_i \in \prod.
\end{equation}

The integration of these concepts within the OPUS architecture is illustrated in Figure~\ref{fig:opus_detailed_diagram}. OPUS comprises four main components: the Interface, LLM, User, and Robotic System. Initially, the Interface transforms raw sensory inputs (such as visual data from the robotic system) into textual observations suitable for the LLM through specialized parsing mechanisms, including vision models that generate object labels. These observations ($O$) are then combined with the user's request ($R$) and available API keywords ($T$) to form a comprehensive prompt for the LLM.

The LLM processes this prompt, reasoning through the task, and produces a sequence of textual API keywords corresponding to feasible robotic commands. The Interface subsequently parses these keywords and translates them into executable robotic actions, represented as policies $\{\pi_i\}_{i=1}^n$, which are then executed by the robotic system.

\subsection{Prompt Construction}
\label{sec:prompt}

\begin{figure}[ht!]
    \includegraphics[width=\linewidth]{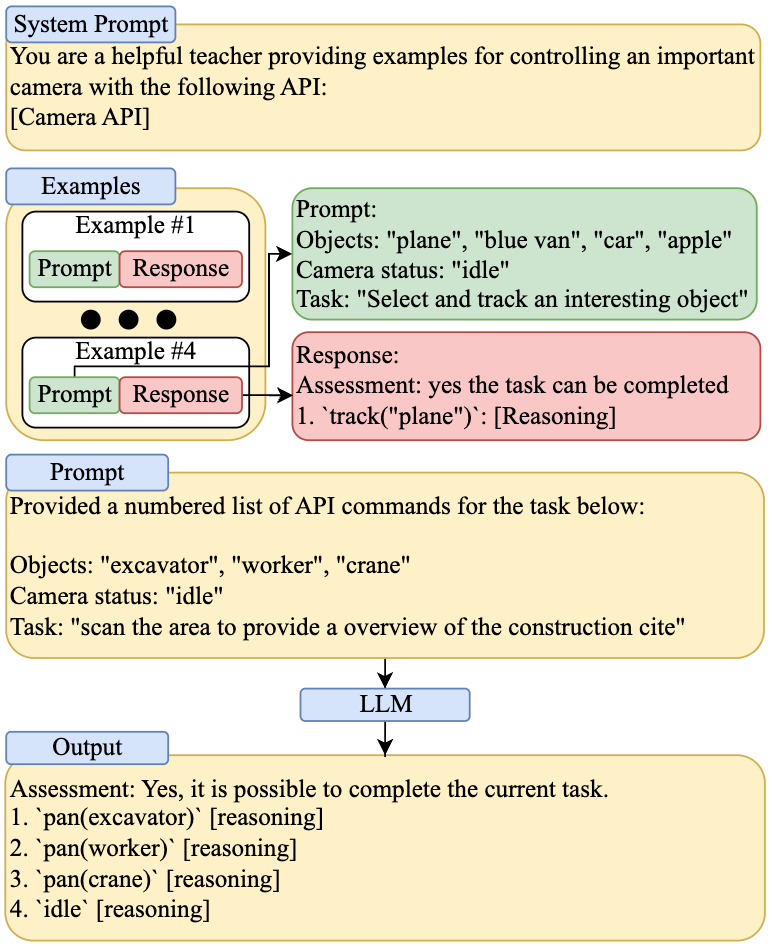}
    \caption{PTZ Prompt Template: The system prompt includes the high-level API, grounding the LLM in the robotic system by describing system capabilities and providing optional task examples. Each example illustrates User-LLM interactions via prompt-response pairs ($p$, $r$). The prompt concludes with the robotic system's state, observations ($O$), and the user request ($R$). The LLM assesses the relevance of $R$ in the context of $O$ and $T$, subsequently generating appropriate API commands to accomplish the requested task.}
    \label{fig:PTZ_prompt_template}
\end{figure}

As illustrated in Figure~\ref{fig:PTZ_prompt_template}, the prompt generated for the LLM integrates several clearly defined components. Initially, the prompt provides a concise robotic system description, clarifying the operational scope and explicitly stating the LLM's role as an assistant that generates executable robotic commands. Subsequently, the prompt presents the Camera API, comprising API keywords and detailed descriptions of associated policies ($T$). For instance, policies related to camera zooming, such as \verb|zoom(value)|, clearly specify valid parameter ranges, including minimum and maximum zoom levels.

Next, the prompt offers illustrative examples demonstrating successful LLM-generated command sequences for similar user requests. Finally, the prompt details current observations ($O$) of the robotic environment and explicitly states the user's request ($R$), which may range from abstract instructions to precise commands referencing specific API keywords. By explicitly incorporating $T$, $O$, and $R$ into the prompt, the LLM receives structured and complete context, enabling accurate and effective generation of robotic commands.

\subsection{Training Dataset Generation}

To generate  a synthetic dataset, $G_s$, essential for SFT, we took inspiration from the Self Instruct studies \citep{wang2023selfinstruct,taori2023alpaca}. Our process begins with the creation of a manually curated seed dataset $G_m$, each element of which, $(p, r)_i$, comprises a conversation containing one input prompt $p$ and one corresponding output response $r$. The prompt $p$ includes a task, relevant observations (e.g., visible objects), and the state of the robotic system, while the response $r$ is composed of an ordered collection of API keywords that directly address the prompt's requirements. Each conversation is further contextualized within a specific setting, such as "construction" or "industrial", to simulate varied environmental conditions.

To generate $G_s$,  we employ a generation prompt that incorporates 8 example $(p, r)$ conversations, with 4 sampled randomly from $G_m$ to anchor the generation in real-world examples and 4 from $G_s$ to foster creative expansion. The prompt also specifies an environmental keyword to define the types of objects present in the camera's view, such as "construction" or "urban". The style of generated conversations is guided by keywords like "panning", "unique", and "creative", which dictate the LLM's output style. This strategy not only ensures the diversity of the dataset but also enhances the practical utility of the LLM in varying scenarios.

Each generation prompt is designed to produce multiple conversations to optimize generation efficiency and reduce costs. For instance, one prompt may request \textit{10 "creative" instances}, while another may seek \textit{10 "panning" instances}. This batched generation approach efficiently leverages Gemini Pro, freely available at the time of writing, to populate $G_m$ with 50,000 PTZ camera examples.

After generating $G_m$, we filter out low-quality instances. Low-quality instances are automatically identified and filtered out by the Interface that extracts generated API keywords from the LLM’s responses. The parser checks these keywords against the robotic API and objects detected in the environmental context. Instances where the parser fails to identify a coherent sequence of API keywords or detects an invalid keyword are excluded from the dataset. 

This filtering method eliminates approximately 30\% of generated instances, resulting in a refined, high-quality dataset of approximately 36,000 conversations. This method ensures that our dataset not only supports effective training but also aligns closely with realistic operational scenarios.

\subsection{Test Dataset Generation}


Before running any experiments or fine-tuning models, we create a test dataset, $G_t$, consisting of 100 conversations. To do this, we generate samples using the procedure above, but with a human in the loop. Each generated response is modified and validated for correctness by the human, and added to $G_t$. Throughout this process, we specifically focus on selecting more challenging examples and strive to diversify the collected examples to enhance $G_t$'s robustness. Additionally, we perform manual refinements to enrich $G_t$, making it more complex and nuanced compared to the training set. This ensures $G_t$ effectively evaluates the models' performance across a broader and more challenging range of scenarios. We repeat this selection and enhancement process until $G_t$ is complete.

\subsection{Model Fine-tuning and Training Setup}

Utilizing $G_s$, we fine-tuned the Zephyr-7B-Beta model\citep{tunstall2023zephyr}, employing the training framework provided by Hugging Face \citep{wolf2019huggingface}. 
To enhance efficiency and optimize training, we implemented Parameter-Efficient Finetuning (PEFT) \citep{xu2023parameterefficient} and Quantized Low Rank Adaptation (QLoRA) \citep{dettmers2023qlora}. The training was conducted using an AdamW optimizer \citep{loshchilov2019decoupled} with a learning rate of $2.5 \cdot 10^{-5}$. Training was terminated after 900 steps when minimal changes in the loss function were observed, indicative of convergence. The training was completed in approximately 4 hours using 2 80GB A100 GPUs. The fine-tuning protocol established for Zephyr-7B-Beta was also applied to other open-source models, including Tiny-Llama-1.1B \citep{zhang2024tinyllama} and Mistral-7B-Instruct \citep{jiang2023mistral}, ensuring a consistent approach across different architectures. The manually curated test dataset ($G_t$) thus serves as a high-quality benchmark that is more challenging than the training data, allowing us to rigorously evaluate model performance under demanding conditions.

\subsection{Simulation and Evaluation Metrics}

To systematically evaluate model performance, we developed a simulator ($\Gamma$) capable of benchmarking commands from both the LLM and expert operators. The simulator maps an ordered set of API commands $\lbrace y_i \rbrace{i=1}^{C}$ into a sequence of visual frames $\lbrace f_i \rbrace_{i=1}^{N}$, defined as $\Gamma(\lbrace y_i \rbrace_{i=1}^C) = \lbrace f_i \rbrace_{i=1}^N$. To ensure consistent evaluation, objects and environments within the simulator remain static across runs.

BMA calculates the average Intersection-over-Union (IOU) across all corresponding frames, evaluating temporal and spatial accuracy:

\begin{equation}
\textstyle
\label{box_match_accuracy}
\textbf{BMA} = \sum_{i=1}^{\text{max}(N_{LLM},N_{E})} \text{IOU}(\hat{f}_{\text{min}(i,N_{LLM})}, \; f_{\text{min}(i,N_{E})})
\end{equation}

Area Accuracy (AA) evaluates spatial coverage over the entire scenario by computing the IOU of all areas visited by the cameras:

\begin{equation}
\textstyle
\textbf{AA} = IOU( \bigcup_{i = 1}^{N_{LLM}} \hat{f}_i,\;\; \bigcup_{i = 1}^{N_E} f_i)
\label{area_accuracy}
\end{equation}

Intuitively, BMA rewards accuracy in both timing and spatial targeting; measuring how well the LLM's commands match the expert's commands on a frame-by-frame basis, capturing both spatial accuracy and timing. Whereas AA measures holistic spatial coverage regardless of timing; evaluating overall spatial coverage regardless of timing, making it a more forgiving metric that focuses on whether the right areas were eventually covered.


\section{Results}
\label{sec:results}

\begin{table}[ht!]
\centering
  \caption{Comparison of PTZ camera task performance across various open-source and closed-source models. The fine-tuned Zephyr-7B (OPUS) achieves the highest AA score and comparable BMA performance despite having significantly fewer parameters.}
  \begin{tabular}{lcccc}
    \toprule
    Model&  \# Parameters & BMA & AA\\
    \midrule
    T-Llama \citep{zhang2024tinyllama} &1.1B& 0.20 & 0.40\\
    Mistral \citep{jiang2023mistral} &7B& 0.34& 0.49\\
    Zephyr \citep{tunstall2023zephyr} &7B& 0.52& 0.70\\
    Mixtral \citep{jiang2024mixtral} &45B& 0.55& 0.73\\
    LLama-2 \citep{touvron2023llama}  &70B& 0.35& 0.68\\
    LLama-3 \citep{dubey2024llama} &70B& 0.61& 0.82\\
    GPT-3.5 \citep{brown2020language} & $\sim$175B & \textbf{0.73}& 0.86\\
    Gemini Pro \citep{geminiteam2024gemini} &\textgreater200B & 0.59& 0.77\\
    GPT-4 \citep{openai2024gpt4} &$\sim$1.8T& 0.72& 0.83\\
    Zephyr - OPUS Trained &7B& 0.71& \textbf{0.89}\\
  \bottomrule
\end{tabular}
\label{tab:sft_results}
\end{table}

To evaluate OPUS's effectiveness, we benchmarked its performance against both closed-source models (e.g., Gemini Pro \citep{geminiteam2024gemini}, GPT-3.5 \citep{brown2020language}, GPT-4 \citep{openai2024gpt4}) and open-source alternatives (e.g., T-Llama \citep{zhang2024tinyllama}, Mistral \citep{jiang2023mistral}, Zephyr \citep{tunstall2023zephyr}). The results, summarized in \Cref{tab:sft_results}, demonstrate that our fine-tuned Zephyr – OPUS Trained (Zephyr-7B-OPUS) achieves the highest AA metric (0.89), surpassing even much larger models such as GPT-4 and LLama-2. In terms of BMA, Zephyr – OPUS Trained achieves comparable performance (0.71), slightly outperforming Gemini Pro (0.59) despite having significantly fewer parameters.

\begin{table}[ht!]
\centering
  \caption{SFT Results: Comparison between base and fine-tuned model across zero-shot and multi-shot prompting}
  \begin{tabular}{lcccc}
    \toprule
    Model& Finetuned& Prompting &BMA&AA\\
    \midrule
    T-Llama-1.1B & no& multi-shot&0.20 & 0.40\\
    T-Llama-1.1B & yes& zero-shot&0.25&\textbf{0.67}  \\
    T-Llama-1.1B & yes& multi-shot&\textbf{0.32} &0.55 \\
    \\
    Mistral-7B & no& multi-shot&0.34 & 0.49\\
    Mistral-7B & yes& zero-shot& 0.35 & 0.73\\
    Mistral-7B & yes& multi-shot&\textbf{0.60} & \textbf{0.78}\\

    \\
    Zephyr-7B & no& multi-shot&0.52 & 0.70\\
    Zephyr-7B & yes& zero-shot&\textbf{0.71} & \textbf{0.89}\\
    Zephyr-7B& yes& multi-shot&0.63 & 0.79\\

  \bottomrule
\end{tabular}
\label{tab:SFT_results}
\end{table}

To further analyze the effectiveness of our fine-tuning approach, we compared the fine-tuned models against their base models across zero-shot and multi-shot prompting scenarios (\Cref{tab:SFT_results}). The results clearly demonstrate that SFT significantly improves performance, with all fine-tuned models consistently outperforming their base counterparts. Notably, Zephyr-7B demonstrates the largest gain, improving from a base BMA of 0.52 and AA of 0.70 in multi-shot prompting to 0.71 (BMA) and 0.89 (AA). This improvement underscores the substantial benefit of fine-tuning on domain-specific tasks.


\begin{table}[ht!]
    \centering
  \caption{Comparison of prompting techniques based on BMA and AA metrics. The fine-tuned Zephyr – OPUS Trained model significantly outperforms alternative prompting methods applied to Gemini Pro.}
  \label{tab:freq}
  \begin{tabular}{llcc}
    \toprule
    Model&Prompting Technique&BMA&AA\\
    \midrule
    Gemini-Pro & Single-Shot&0.40 & 0.67\\
    Gemini-Pro & Multi-shot&0.59 & 0.77\\
    Gemini-Pro & Reflexion \citep{shinn2023reflexion} &0.57 & 0.75\\
    Gemini-Pro & High-Medium Planner \citep{sermanet2023robovqa} &0.34 & 0.74\\
    Zephyr - OPUS Trained & Finetuned &\textbf{0.71} & \textbf{0.89}\\
  \bottomrule
\end{tabular}
\label{tab:opus_prompting_results}
\end{table}

\Cref{tab:opus_prompting_results} compares the fine-tuned Zephyr – OPUS Trained model with its larger parent model, Gemini Pro, which was used to generate the synthetic dataset. Remarkably, Zephyr – OPUS Trained surpasses Gemini Pro’s performance by over 20\% in BMA and approximately 15\% in AA, despite being around 28 times smaller. Furthermore, Zephyr – OPUS Trained outperforms more complex prompting techniques such as Reflexion \citep{shinn2023reflexion} and High-Medium Planner \citep{sermanet2023robovqa}, which require iterative prompting with Gemini Pro. Thus, the fine-tuned Zephyr model not only achieves superior performance but also significantly reduces computational overhead by relying on a single prompt, underscoring its efficiency and practicality for real-world deployment.

\subsubsection*{Statistical Analysis}

We conducted a bootstrap resampling statistical analysis to rigorously evaluate the performance of the Fine-Tuned Zephyr model relative to its baseline and larger counterparts. This analysis involved 1,000 iterations, each time randomly sampling 100 data points from the test dataset with replacement. At a significance level of \( p = 0.05 \), the Fine-Tuned Zephyr model demonstrated statistically significant improvements in both AA and BMA metrics compared to Gemini Pro, GPT-4, and the base Zephyr model. Furthermore, it outperformed GPT-3.5 across both metrics, with statistically significant improvements specifically observed in the BMA metric. These results reinforce the robustness and effectiveness of our fine-tuning approach.

\section{Conclusion and Future Directions}

In this work, we introduced OPUS, a framework designed to enable efficient and robust control of PTZ cameras via natural language. By employing a high-level API and SFT on a synthetic dataset, OPUS achieves performance comparable to substantially larger models, such as GPT-4, while remaining efficient enough for practical, real-time deployment. The framework's compact size and inference efficiency eliminate the need for computationally expensive hardware, significantly enhancing its practicality for deployment in resource-constrained environments.

Despite these promising results, our approach has several limitations. Although the real-world demonstration presented in the \Cref{appendix} provides an initial indication of OPUS’s potential, reliance on a synthetic dataset still limits the system’s exposure to the full range of complexities encountered in practical scenarios. In particular, while our synthetic dataset incorporates varied contexts and challenging tasks, it cannot fully replicate dynamic lighting variations, sensor noise, and other intricate sensory conditions present in real environments. More extensive testing and evaluation in these complex settings are required to further refine system robustness and reliability. Future work will focus on augmenting the synthetic dataset with diverse real-world examples, integrating richer sensory inputs, and validating model performance through deployment on physical robotic systems.

In addition, incorporating a human-in-the-loop approach could significantly enhance OPUS’s practical performance. Users could actively correct command sequences, rate system effectiveness, or provide natural language feedback. Such continuous feedback mechanisms would enable the model to iteratively learn from realistic user interactions, further refining its capabilities without sacrificing inference efficiency. Exploring these interactive training strategies represents a promising direction for future research.

Beyond PTZ camera control, OPUS demonstrates potential for broader robotics applications. The framework could readily extend to other robotic systems requiring precise natural-language-driven commands. Potential applications include robotic manipulators, mobile robots, or drones, each with their unique temporal dynamics and operational constraints. Adapting OPUS to these contexts would involve incorporating task-specific APIs, enhancing training datasets with domain-specific examples, and addressing the distinct challenges posed by dynamic and uncertain environments. Such expansions represent significant opportunities for advancing the applicability and generality of language-driven robotic control systems.


\appendix
\label{appendix}
\setcounter{figure}{0}

\captionsetup[figure]{
  name=Appendix Fig.,    
  labelformat=simple,    
}
\renewcommand{\thefigure}{\arabic{figure}}

\captionsetup[figure]{name=Appendix Fig,labelsep=period}
\renewcommand{\thefigure}{\arabic{figure}}

\crefname{figure}{Appendix Fig.}{Appendix Figs.}
\Crefname{figure}{Appendix Fig.}{Appendix Figs.}

\begin{figure*}[!b] 
    \centering
    \includegraphics[width=.95\linewidth]{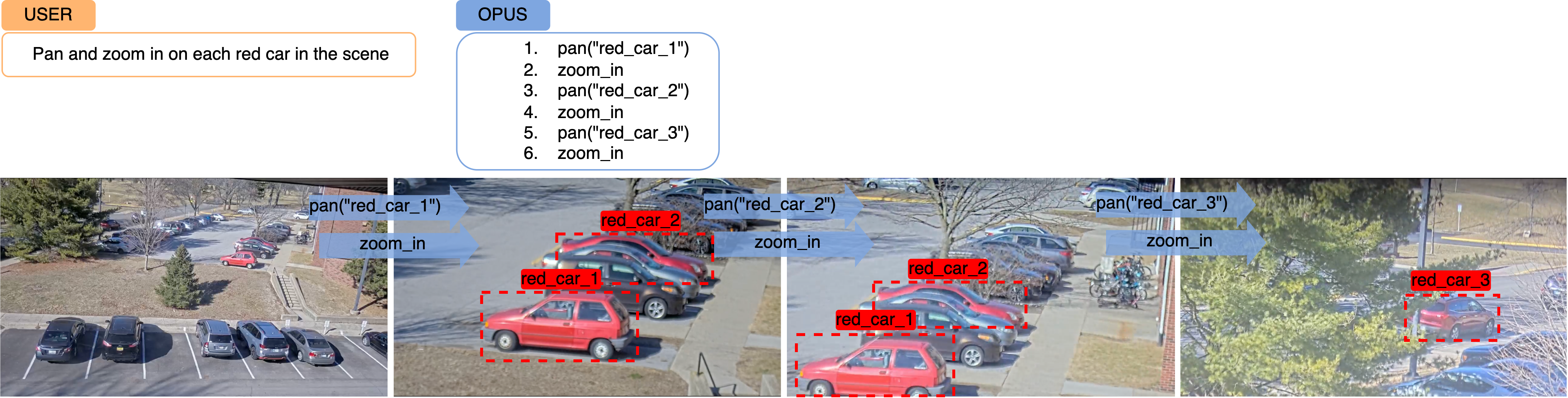}
    \caption{Demonstration of OPUS in a real-life setting. OPUS is given an open-ended task, in this case to zoom in on each red car in a crowded parking lot.
    After the OPUS Interface identifies objects of interest, these objects are 
    passed to a LLM which generates a sequence of high-level API instructions.
    Then these commands are parsed by the OPUS Interface, and executed on the PTZ
    Camera.}
    \label{fig:opus_irl}
\end{figure*}

\begin{figure}[H]
\centering
    \includegraphics[width=.95\linewidth]{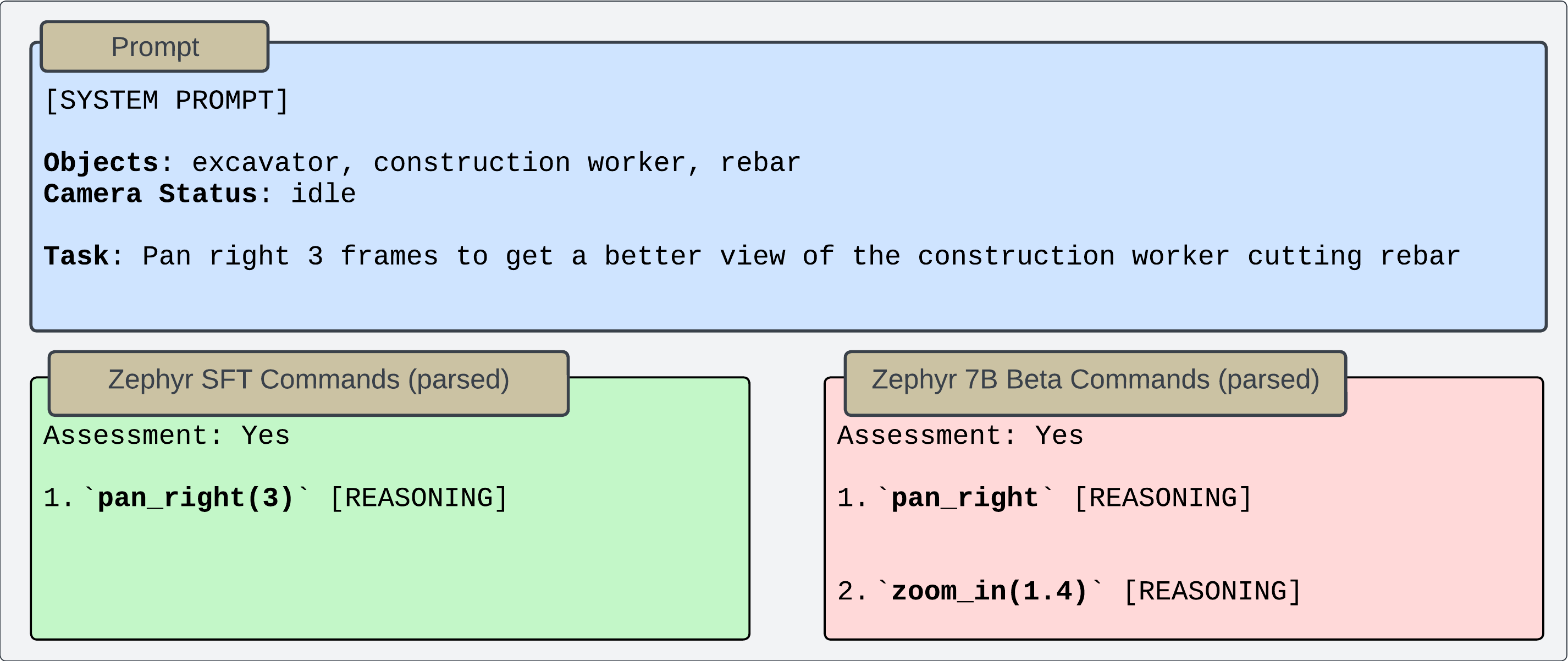}
    \caption{Example response from Zephyr SFT and Zephyr 7B Beta for a task in a mining environment. Zephyr SFT correctly parses the user prompt, panning by only 3 frames to the right. Zephyr 7B Beta instead pans all the way to the right calling an unrelated command.}
\label{fig:ex1}
\end{figure}

\Cref{fig:opus_irl} illustrates the OPUS system in action, showcasing its capability to autonomously identify and focus on specific targets within a complex real-world environment. This demonstration provides tangible evidence of the system's practical application and its effective integration of high-level command parsing with PTZ camera control.

\section{Camera API / Full System Prompt:}


\section{Response Comparison}

\begin{figure}[H]
\centering
    \includegraphics[width=.95\linewidth]{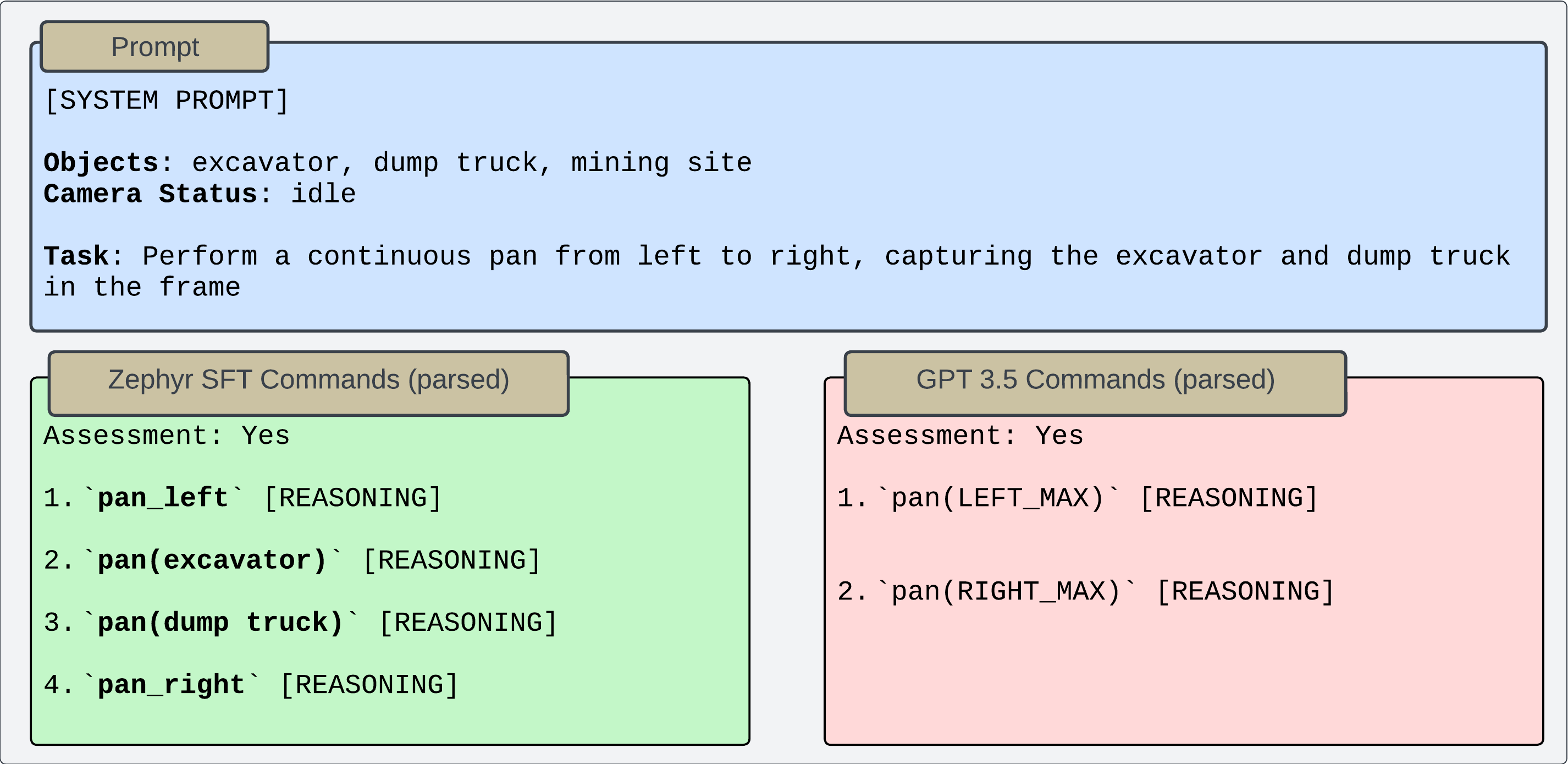}
    \caption{Example response from Zephyr SFT and GPT-3.5 for a task in a construction environment. Zephyr SFT correctly realizes that the user requested to capture the excavator and dump truck while panning from the left, while GPT fails to do so.}
\label{fig:ex2}
\end{figure}

In \Cref{fig:ex1} and \Cref{fig:ex2}, we present examples of common pitfalls of LLMs that were removed through SFT. Smaller LLMs suffered from using incorrect commands, while occasionally hallucinating non-existed commands and objects that couldn't be parsed by the Robotic-LLM Interface. Larger LLMs would use correct commands, but suffered from generating sub-optimal responses, frequently failing to complete more complicated user requests.

\bibliographystyle{IEEEtranN}
\bibliography{references.bib}

\end{document}